# A Novel Region Duplication Detection Algorithm based on Hybrid Approach

Ms. Kshipra A. Tatkare[a,] Dr. Manoj Devare[b] *

[a]Reaserch Scholar, Amity University, Mumbai, India
[b]Associate Professor, Amity University Mumbai, India

**Abstract**

The digital images from various sources are ubiquitous due to easy availability of high bandwidth Internet. Digital images are easy to tamper with good or bad intentions. Non-availability of pre-embedded information in digital images makes the tampering detection process more difficult in case of digital forensics. Thus, passive image tampering is difficult to detect. There are various algorithms available for detecting image tampering. However, these algorithms have some drawbacks, due to which all types of tampering cannot be detected. In this paper researchers intend to present the types of image tampering and its detection techniques with example based approach. This paper also illustrates insights into the various existing algorithms and tries to find out efficient algorithm out of them.

**© 2019 – Authors.**

## 1. Introduction

The first incidence is occurred in decade of 1840 according to the image tampering history. The first forged image has been produced by Hippolyta Bayard. His famous tampered picture is committing suicide. There is hardly difference between Digital Image Tampering and Conventional Image Tampering. The only difference in the digital images and conventional images is black films and photographs. Nowadays, anyone can swimmingly produce fake or forged image with the help of image editing applications like GIMP, Adobe Photoshop, Picasa, lightroom etc.

### 1.1 Objectives

The alteration in digital image is the main aim of image tampering. This image tampering may be done to hide the information in an image or purposely to send some private or confidential signals from sender to receiver. The objective of this paper is to build a system, which will detect all the types of tampering in an image.

*1.2 Applications*

Image tamper detection techniques can be applied in following areas to detect tampering:
1.    Surveillance System
2.    Medical Imaging
3.    Journalism
4.    Criminal Investigation
5.    Social Networking sites threats
6.    Intelligence Services

The leftover section of the paper is structured as: Literature review is discussed in Section II. Section III presents Results and Evaluation parameters values of different author's implemented algorithms. Section IV addresses the Analysis of exiting techniques. Section V concludes with future scope.

## 2. Literature Overview

*2.1 Image Tamper Detection*

Image tamper detection is a part of Digital Forensic, which has an objective of revitalization with analysis in tampered images [1]. Classification of image tampering is shown in Fig. 1. There are two approaches for image tampering which are Active and Passive.

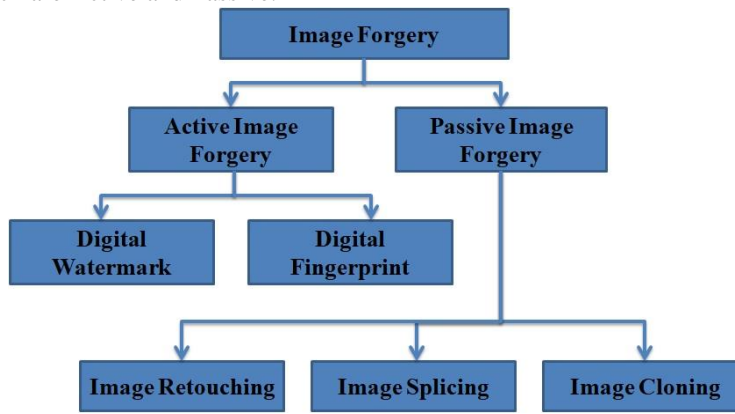

Fig.1. Classification of Image Tampering

In active Image Tampering, an image is having pre-embedded information, such as fingerprint or watermark. Original image knowledge is required to detect tampering in this type of techniques [2].

In passive image tampering, pre-embedded information of an image is not required [3]. Knowledge of an Original image is also not required in this type of tamper detection technique [4]. Using Passive tamper detection techniques, the tampering can be detected for major images from magazines, newspapers, posters, various websites etc [5].

There are three categories of passive image tampering, which are image retouching, image cloning and image splicing. If an image is tampered using Blurring, Scaling and Rotation, then it is an example of an image retouching. An image retouching is also done using Pixel temperature change and object colour modifications. Fig. 2 shows the pattern of this type of tampering.

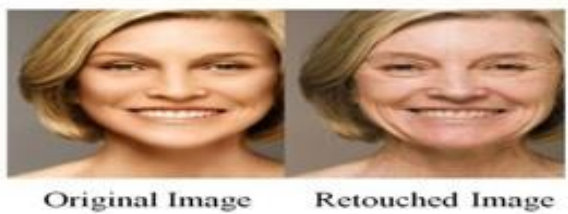

Fig.2. Example image of a typical Re-touching

In an image cloning technique, the duplication is done in an image using one or more pieces of the same image [6]. Hence, it is also called as region duplication. Fig.3 shows the pattern of this type of tampering and its detection.

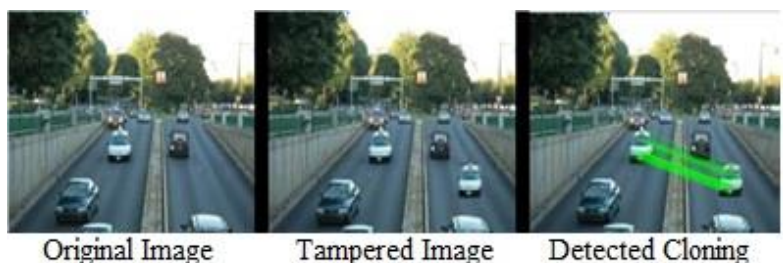

Fig.3. Example of Example image of a typical image cloning

In the third type of tampering technique i.e. image splicing, one or more pieces from different images are combined together. Fig. 4 shows pattern of this type of tampering.

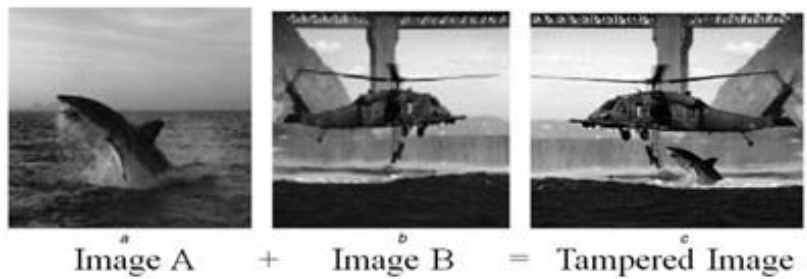

Fig.4. Example image of a typical Image Splicing

Image tamper detection, in passive techniques, is not an easy job as compared to the active techniques. It is because in passive techniques pre-embedding information is not present [7]. So, the focus of this study is on passive image tamper detection.

*2.2 The Processing Pipeline for Detection of Passive Image Tampering*

The process of passive image tamper detection includes various steps like, essential pre-processing techniques for an image, Feature can be extracted from an image, matching of feature vectors, filtering as per the requirement and post-processing [8]. There is a general processing pipeline to detect tampering in an image which is shown in Fig.5.

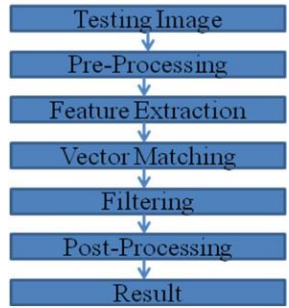

Fig.5. Process of Image Tamper Detection

Pre-processing technique on an image includes colour to gray scale translation, image resizing, conversion to binary format , Skeletonisation, Morphological Operation, and separation of plane [9][10].

Feature extraction is known as dimensional reduction, which is classified into block based and key point based [8]. The focus is on high entropy points of an image in key-point approach. The various algorithms are applied on those high entropy points to find out feature vectors [11]. Similarly, the image splits into blocks of specified size like 8*8 or 16*16 and then feature extraction algorithms are applied [12]. Feature vectors of an image can be obtained using various algorithms which are discussed in later section.

The next step is the matching of feature vectors. For matching process, there are various techniques like kd-tree algorithm, Lexicographic sorting etc. Thus, these matching pairs are of tampered part of an image.

False matches can be minimized using filtering. False tampering is detected using intensities of neighbouring pixels. Matched pairs are conserved using final step post-processing. The processing pipeline for detection of an image tampering is shown in Fig.5.

Major important algorithms are reviewed as follows:

**Scale Invariant Feature Transform (SIFT)**

Features of digital images are extracted using SIFT algorithm. The extracted features are then compared with each other to trace the duplicate region [13]. The drawback of this method is high dimensionality of the SIFT descriptors. There is scope for improvement in the existing technique.

**Speed Up Robust Features (SURF)**

From digital images, features are extracted using SURF algorithm and then using adjacent neighbouring; the matching of vectors is done. SURF features are invariant to scale and rotation. It is also detects tampering if there are noise, blurring and jpeg compression present.

**Zernike**

ZERNIKE moments are superior in all the terms with respect to insensitivity to image noise, information content etc [14]. These moments are rotation invariant [15]. Affine transformation is still weak in this method.

**Hue Moments**

These moments are rotation invariant. To identify various typed characters, Hue's moments are used in a pattern recognition research. The computation of Hue moment from normalized and centralized moments having degree three are shown in nth Hue invariant moment:

$$I1 = \eta_{20} + \eta_{02} \tag{1}$$

$$I2 = (\eta_{20} - \eta_{02})^2 + 4\eta^2_{11} \tag{2}$$

$$I3 = (\eta_{30} - 3\eta_{12})^2 + (3\eta_{21} - \eta_{03})^2 \tag{3}$$

$$I4 = (\eta_{30} + \eta_{12})^2 + (3\eta_{21} + \eta_{03})^2 \tag{4}$$

$$I5 = (\eta_{30}-3\eta_{12})(\eta_{30}+\eta_{12})[(\eta_{30} + \eta_{12})^2- 3(\eta_{21}+\eta_{03})^2]+(3\eta_{21}-\eta_{03})(\eta_{21}+\eta_{03})[3(\eta_{30}+\eta_{12})^2-(\eta_{21}+\eta_{03})^2] \tag{5}$$

$$I6 = (\eta_{20} - \eta_{02}) [(\eta_{30} + \eta_{12})^2 - (\eta_{21} + \eta_{03})^2 (42) +4\eta_{11}(\eta_{30} + \eta_{12}) (\eta_{21} + \eta_{03})] \tag{6}$$

To distinguish mirror image, a skew invariant is used as follows:

$$I7 = (3\eta_{21}-\eta_{03})(\eta_{30}+\eta_{12})[(\eta_{30}+\eta_{12})^2-3(\eta_{21}+\eta_{03})^2]+(\eta_{30}-3\eta_{12})(\eta_{21}+\eta_{03})[3(\eta_{30}+\eta_{12})^2-(\eta_{21}+\eta_{03})^2] \tag{7}$$

In this paper, a hybrid or mixed proposition, based on key-point and blocks, is presented with experimental results. This approach elevates the drawbacks of traditional approaches. The memory requirement is higher for algorithms, which are based on key-points than based on blocks. The false positive detection in image tampering is occurred due to JPEG compression in block-based approach [16]. Amount of high-dissimilarity and identity resemblance of non-copied regions are more challenging tasks in key-point based algorithms, whereas block based approach is efficient to detect the same. In general, the key-point based algorithms give best results for an image, which is rough in nature; whereas block based algorithms give best results for an image, which is smooth in nature. These are benefits and drawbacks of traditional approaches of image tamper detection. Hence, the presented approach elevates the drawbacks of traditional approaches. In the presented hybrid approach, global features are extracted using Hue Moments and local features are extracted using SIFT algorithm [17] [18].

## 3. Results and Evaluation Parameters

There are various descriptors explained in literature above. For this experiment MICCF220 database is used. For this experiment, a novel combined approach based on key-points (SIFT) and blocks (Hue Moments) are used. The results of various scenarios are shown in the following figures which are Fig. 6, Fig. 7 and Fig. 8:

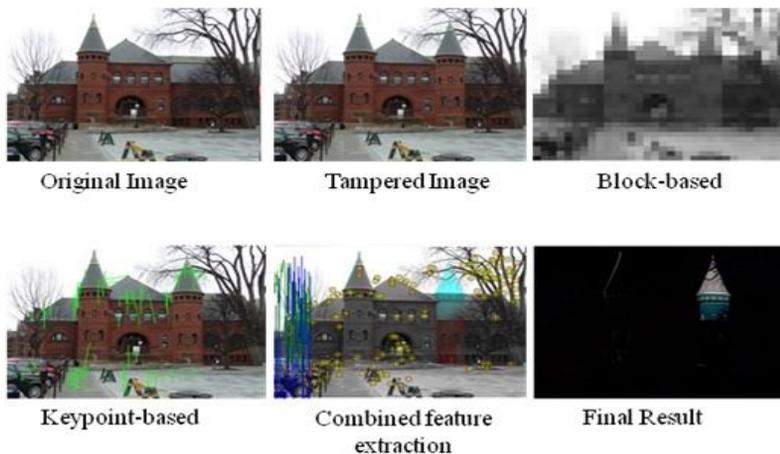

Fig.6. Result of hybrid (key-point based with block based) approach

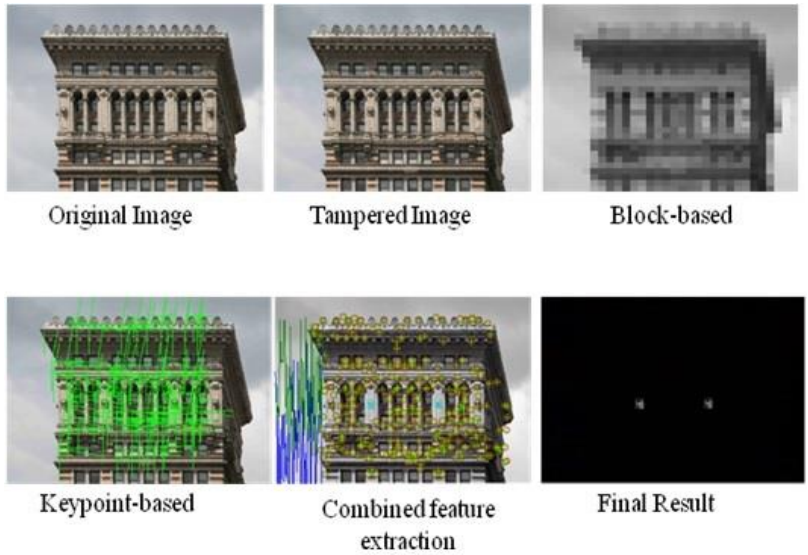

Fig.7. Result of hybrid system with Rough Surface

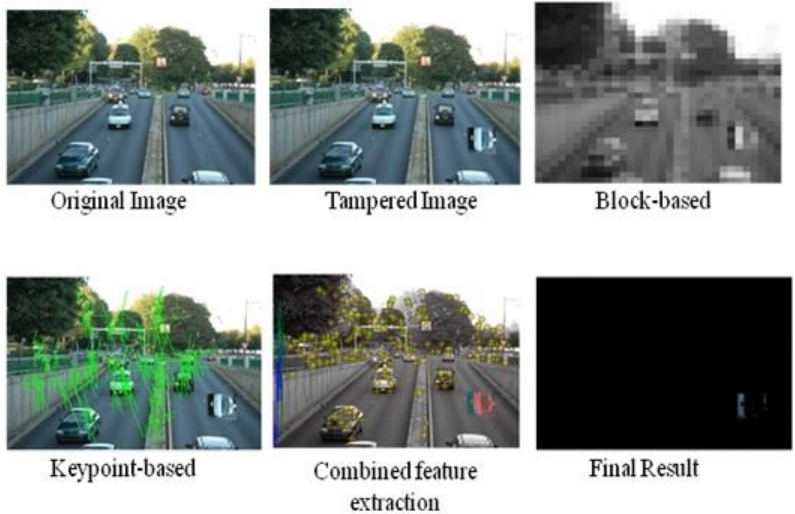

Fig.8. Result of hybrid system with rotated pasted region

In above experimental results, top row contains original image, tampered image and images divided into overlapping blocks sequentially and bottom left window shows extracted key-points, bottom middle window shows combined features of key-point and block based approach and bottom right window shows detected tampered region of image.

The evaluation parameters for above experimental result are Recall, Precision, F1 Score and Accuracy.

## 4. Comparative Analysis

Precision can be calculated as the possibility that detected tampering is of true tampering. Recall can be calculated as the possibility that, detected image is tampered one. A function of True Positive (TP) can be computed, which is F1-score. The computational formulas of Precision, Recall, F1 score and Accuracy are:

*Precision (P) = (True Positive (TP) / True Positive (TP)) + False Positive (FP)*         *(8)*

*Recall (R) = (True Positive (TP) / True Positive (TP)) + False Negative (FN)*         *(9)*

*Accuracy = (TP+ True Negative (TN) / (TP+FN+TN+FP)*         *(10)*

*F1 score = 2(P\*R)/ (P+R)*         *(11)*

The comparative analysis of these results is mentioned in table 1.

Table 1  Comparative Analysis of Hybrid System

| Method | Precision | Recall | Accuracy | F1 Score |
|---|---|---|---|---|
| Hue [2] | 67.61 | 74.89 | 70.14 | 80.67 |
| SIFT [2] | 88.37 | 79.17 | 67.18 | 90.53 |
| Hybrid Approach | 97.68 | 96.04 | 92.48 | 96.85 |

The comparison is shown using graphical presentation in fig.9. The hybrid system, which is the combination of SIFT and Hue Moments, gives the highest values for all analytical parameters in regards to detect tampering.

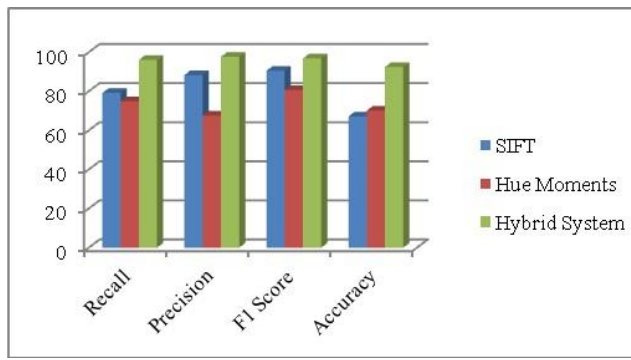

Fig.9. Comparative Analytical graph of SIFT, Hue and Hybrid Approach

## 5. Conclusion

This study concludes that; the hybrid approach elevates the drawbacks of traditional algorithms by combining feature vectors of SIFT and Hue Moments algorithms. As results show hybrid approach in image tamper detection are the most efficient in all analytical parameters that are Recall, Precision, F1 Score and Accuracy. As it has almost 95% significance level, it is the most appropriate method for different genres of images such as flat surface, rough surface, multiple tampering in one image, rotated and scaled tampered image. This paper definitely helps other researchers and the domain experts to select image tamper detection techniques and hybrid approach over other techniques and approaches to get efficient results.

## 6. Future Scope

In existing literature, there is no single system present, which can detect the tampering in an image with all three categories of passive approach. If such kind of system will be developed in future, then that will be helpful for social cause.

## References

[1]    Shan Jia, Zhengquan XU, Hao Wang, Chunhui Feng, Tao Wang "Coarse-to-Fine Copy-Move Forgery Detection for Video Forensics",IEEE Translations and content mining, Volume 6, Pages 25323-25335, May 2018.

[2]    Vincent Christlein, Christian Riess, Johannes Jordan, Corinna Riess, Elli Angelopoulou, "An Evaluation of Popular Copy-Move Forgery Detection Approaches", Proceedings of the IEEE Transactions on Information Forensics and Security, pages 1 - 26, November 2012 .

[3]    Wei Wang, Jing Dong, and Tieniu Tan, "A Survey of Passive Image Tampering Detection",Springer-Verlag Berlin Heidelberg, LNCS 5703, Pages 308322, 2009.

[4]    Osamah M. Al-Qershi and Khoo Bee Ee, "Passive Detection of Copy-Move Forgery in Digital Images: State-of-the-art", Forensic Science International Conference, Volume 231, Issues 13, Pages 284295, 10 September2013.

[5]    Shivani Thakur, Ramanpreet Kaur, Dr. Raman Chadha, Jasmeet Kaur, A Review Paper on Image Forgery Detection In Image Processing, IOSR Journal of Computer Engineering (IOSR-JCE) e-ISSN: 2278-0661,p-ISSN: 2278-8727, Volume 18, Issue 4, Ver. I (Jul.-Aug. 2016), PP 86-89.

[6]    Yanjun Cao, Tiegang Gao, Li Fan and Qunting Yang, "A robust detection algorithm for copy move forgery in digital images", Forensic Science International Conference, 214, Pages 33 43, 2012.

[7]    Salma Amtullah and Dr. Ajay Koul, "Passive Image Forensic Method to detect Copy Move Forgery in Digital Images", IOSR Journal of Computer Engineering (IOSR-JCE), Volume 16, Issue 2, Ver. XII, Pages 96-104, Mar-Apr. 2014.

[8]    Somayeh Sadeghi, Hamid A. Jalab, and Sajjad Dadkhah,"Efficient Copy-Move Forgery Detection for Digital Images", World Academy of Science, Engineering and Technology, Volume 6, pages 539 - 542, 2012.

[9]    Chi-Man Pun, Xiao-Chen Yuan and Xiu-Li Bi, "Image Forgery Detection Using Adaptive Oversegmentation and Feature Point Matching", IEEE TRANSACTIONS ON INFORMATION FORENSICS AND SECURITY, VOL. 10, NO. 8, 2015.

[10]   CodrutaO. Ancuti, Cosmin Ancuti, Christophe De Vleeschouwer and Philippe Bekaert, "Color Balance and Fusion for Underwater Image Enhancement", IEEE TRANSACTIONS ON IMAGE PROCESSING, VOL. 27, NO. 1, 2018.

[11]   Jian Li, Xiaolong Li, Bin Yang, and Xingming Sun, "Segmentation-Based Image Copy-Move Forgery Detection Scheme", IEEE TRANSACTIONS ON INFORMATION FORENSICS AND SECURITY, VOL. 10, NO. 3, 2015.

[12]   Vivek Kumar Singh and R.C. Tripathi, "Fast and Efficient Region Duplication Detection in Digital Images Using Sub-Blocking Method", International Journal of Advanced Science and Technology, VOL. 35, Pages 93 - 102, October 2011.

[13]   Lichao Su, Cuihua Li, Yuecong Lai and Jianmei Yang "A Fast Forgery Detection Algorithm based on Exponential-Fourier Moments for Video Region Duplication", IEEE Transaction on Multimedia, DOI 10.1109/TMM.2017.2760098.

[14]   Seung-Jin Ryu, Min-Jeong Lee, and Heung-Kyu Lee,"Detection of Copy-Rotate-Move Forgery Using Zernike Moments", Springer-Verlag Berlin Heidelberg, LNCS 637, Pages 51 65, 2010.

[15]   Chandan singh, Ekta Walia, and Neerja Mittal, "Fusion of Zernike Moments and SIFT Features for Improved Face Recognition", International Conference on Recent Advances and Future Trends in InformationTechnology,Pages26–31,2012

[16]   Haodong Li, Weiqi Luo, Xiaoqing Qiu and Jiwu Huang, "Image Forgery Localization via Integrating Tampering Possibility Maps", IEEE TRANSACTIONS ON INFORMATION FORENSICS AND SECURITY, VOL. 12, NO. 5, 2017.

[17]   Cai-Ping Yan and Chi-Man Pun, "Multi-Scale Difference Map Fusion for Tamper Localization Using Binary Ranking Hashing", IEEE TRANSACTIONS ON INFORMATION FORENSICS AND SECURITY, VOL. 12, NO. 9, 2017.

[18]   Wujie Zhou , Lu Yu ,Yang Zhou, Weiwei Qiu, Ming-Wei Wu and Ting Luo, "Local and Global Feature Learning for BlindQuality Evaluation of Screen Contentand Natural Scene Images", IEEE TRANSACTIONS ON IMAGE PROCESSING, VOL. 27, NO. 5, 2018.

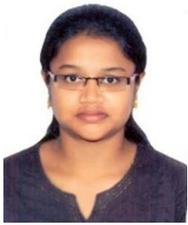

**Ms. Kshipra A**. **Tatkare** received the B.E. degree in Information Technology Engineering from Ramrao Adik Institute of Technology, Nerul, Navi Mumbai, in 2011 and the M.E. degree in Computer Science Engineering from Ramrao Adik Institute of Technology, Nerul, Navi Mumbai, in 2015. She is currently pursuing the Ph.D. degree in Computer Science and Engineering at Amity University, Mumbai, India and Assistant Professor in Ramrao Adik Institute of Technology, Nerul, Navi Mumbai.

From 2011 to 2013, she was working as a PHP Developer. During this period she worked on different Content Management Systems (CMS) like Joomla, PHPFox, Skadate, Dolphin, WordPress, Drupal etc. She got "Employee of the Month" for her best performance. From 2013 to 2015, she was working as a Teaching Assistant while pursuing M.E. degree. She has developed an educational application for her institute. She has published her research paper in 09 national and international conferences and Journals. Her research interest includes Cyber Security, Digital Forensics, Intelligent Systems and Search Engine Optimization.

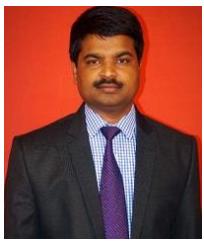

**Dr. Manoj Devare** has completed his education from North Maharashtra University and PhD in Computer Science from Bharati Vidyapeeth University, India. He has served as Post Doctorate Fellow at Centre of Excellence on HPC, Uni. of Calabria, Italy under young researcher scheme. Dr. Manoj is working as Associate Professor at Amity Institute of Information Technology, Amity University Mumbai. He has 16 years of teaching and research experience in Machine Learning and Cloud Computing. His PhD thesis was based on the Congestion Controlling in Homogeneous and Heterogeneous Computer Networks. Dr. Manoj won two "Best Paper Awards", in International Conferences ICSCI 2008, and ICATCSIT 2019. He has been developed Desktop Cloud system during this PDF.

He has 02 Patents pending; and he has edited 06 Scopus indexed book chapters, 06 International Journal Scopus indexed papers, and 15 Conference Papers. He edited two conferences ISBN proceeding/ books. He has delivered 08 national and international keynote speeches. He is working as an Associate Editor of the WOS indexed International Journal of Cognitive Informatics and Natural Intelligence (IJCINI), and International Journal of End-User Computing and Development (IJEUCD). Dr. Manoj is working as an editorial review member of International Journal of Fog Computing (IJFC), Elsevier's Journal of Neuroscience Methods, and Symbiosis International Journal. He is working as a PhD Supervisor in Computer Science.